\documentclass[journal]{IEEEtran}

\usepackage{epsfig}
\usepackage{graphicx}
\usepackage{subfigure}
\usepackage{amsmath}
\usepackage{amssymb}
\usepackage{amsfonts}
\usepackage{booktabs}
\usepackage{xcolor}
\usepackage{cite}

\begin{document}

\title{A Deep Error Correction Network for\\ Compressed Sensing MRI}

\author{Liyan Sun,~ Zhiwen Fan,~ Yue Huang,~ Xinghao Ding,~ John Paisley$^{\dagger}$\\
Fujian   Key   Laboratory   of   Sensing   and   Computing   for   Smart   City,  Xiamen   University, Fujian, China\\
$^{\dagger}$Department of Electrical Engineering, Columbia University, New York, NY, USA
}

\maketitle

\begin{abstract}
Compressed sensing for magnetic resonance imaging (CS-MRI) exploits image sparsity properties to reconstruct MRI from very few Fourier k-space measurements. The goal is to minimize any structural errors in the reconstruction that could have a negative impact on its diagnostic quality. To this end, we propose a deep error correction network (DECN) for CS-MRI. The DECN model consists of three parts, which we refer to as modules: a guide, or template, module, an error correction module, and a data fidelity module. Existing CS-MRI algorithms can serve as the template module for guiding the reconstruction. Using this template as a guide, the error correction module learns a convolutional neural network (CNN) to map the k-space data in a way that adjusts for the reconstruction error of the template image. Our experimental results show the proposed DECN CS-MRI reconstruction framework can considerably improve upon existing inversion algorithms by supplementing with an error-correcting CNN.
\end{abstract}


\IEEEpeerreviewmaketitle

\section{Introduction}

\begin{figure*}[t]
\begin{center}
   \subfigure {\includegraphics[width=1\textwidth]{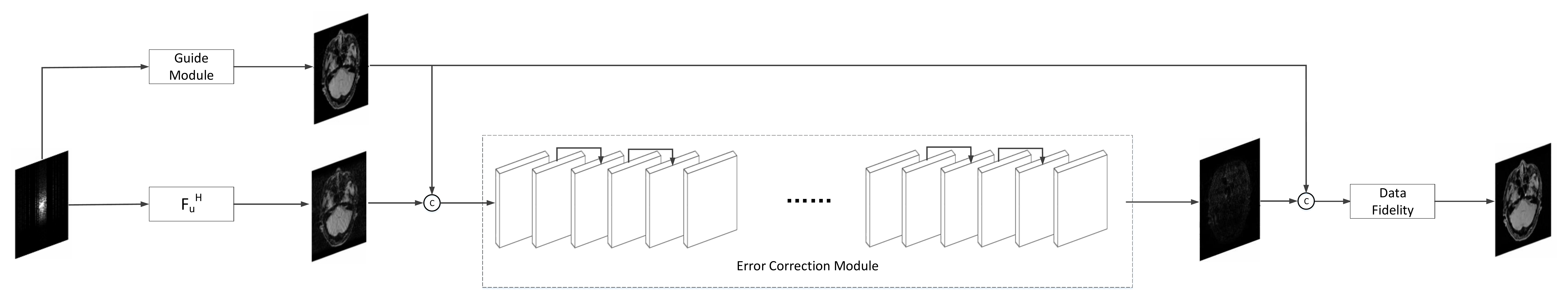}}
   \caption{The proposed Deep Error Correction Network (DECN) architecture consists of three modules: a guide module, an error correction module, and a data fidelity module. The input of the error correction module is the concatenation of the zero-filled compressed MR samples and guidance image while the corresponding training label is the reconstruction error $\Delta x_p$. After the error correction module is trained, the guidance image and feed-forward approximation of the reconstruction error for a test image are used to produce the final reconstructed MRI.}
\label {fig1}
\end{center}
\end{figure*}

\IEEEPARstart{M}{agnetic} resonance imaging (MRI) is an important medical imaging technique, but its slow imaging speed poses a limitation on its widespread application. Compressed sensing (CS) theory \cite{1,2} has been a significant development of the signal acquisition and reconstruction process that has allowed for significant acceleration of MRI. The CS-MRI problem can be formulated as the optimization
\begin{equation}\label{eq1}
\hat x = \mathop {\arg \min }\limits_x \left\| {{F_u}x - y} \right\|_2^2 + \sum\limits_i {{\alpha _i}{\Psi _i}\left( x \right)},
\end{equation}
where $x \in {C^{N \times 1}}$ is the complex-valued MRI to be reconstructed, $F_u\in {C^{M \times N}}$ is the under-sampled Fourier matrix and $y\in {C^{M \times 1}}$ ($M\ll N$) are the k-space data measured by the MRI machine. The first data fidelity term ensures agreement between the Fourier coefficients of the reconstructed image and the measured data, while the second term regularizes the reconstruction to encourage certain image properties such as sparsity in a transform domain.

Recently, deep learning approaches have been introduced for the CS-MRI problem, achieving state-of-the-art performance compared with conventional methods. For example, an end-to-end mapping from input zero-filled MRI to a fully-sampled MRI was trained using the classic CNN model in \cite{16}, or its residual network variant in \cite{17}.
Greater integration of the data fidelity term into the network has resulted in a Deep Cascade CNN (DC-CNN) \cite{18}.
Compared with previous models proposed for CS-MRI inversion, deep learning is able to capture more intricate patterns within the data, which leads to their improved performance.

Recently, the compressed sensing MRI is approved by the Food and Drug Administration (FDA) to two main MRI vendors: GE and Siemens \cite{39}. As the growing needs for application of compressed sensing MRI, improving reconstruction accuracy of the CS-MRI is of great significance. In this paper, we propose a deep learning framework in which an arbitrary CS-MRI inversion algorithm is combined with a deep learning error correction network. The network is trained for a specific inversion algorithm to exploit structural consistencies in the errors they produce. The final reconstruction is found by combining the information from the original algorithm with the error correction of the network.


\section{Related Work}

Much focus of previous work has been on proposing appropriate regularizations that lead to better MRI reconstructions. In the pioneering work of CS-MRI called SparseMRI \cite{3}, this regularization adds an $\ell_1$ penalty on the wavelet coefficients and the total variation of the reconstructed image.
Based on SparseMRI, more efficient optimization methods have been proposed to optimize this objective, such as TVCMRI \cite{4}, RecPF \cite{5} and FCSA \cite{6}. Variations on the wavelet penalty exploit geometric information of MRI, such as PBDW/PBDWS \cite{7,8} and GBRWT \cite{9}, for improved results. Dictionary learning methods \cite{10,11,12,38} have also been applied to CS-MRI reconstruction, as have nonlocal priors such as NLR \cite{13}, PANO \cite{14} and BM3D-MRI \cite{15}. These previous works can be considered sparsity-promoting regularized CS-MRI methods that are optimized using iterative algorithms. They also represent images using simple single layer features that are either predefined (e.g., wavelets) or learned from the data (e.g., dictionary learning).

Previous work has also tried to exploit regularities in the reconstruction error in different ways. In the popular dynamic MRI reconstruction method k-t FOCUSS \cite{21,22}, the original signal is decomposed into a predicted signal and a residual signal. The predicted signal is estimated by temporal averaging, while the highly sparse residual signal has a $l_1$-norm regularization. An iterative feature refinement strategy for CS-MRI was proposed in \cite{23} to exploit the structural error produced in each iteration. In \cite{24}, the k-space measurements are divided into high and low frequency regions and reconstructed separately. In \cite{25} the MR image is decomposed into a smooth layer and a detail layer which are estimated using total variation and wavelet regularization separately.  In \cite{26}, the low frequency information is estimated using parallel imaging techniques. These methods each employ a fixed transform basis.

\section{Problem Formulation}

Exploiting structural regularities in the reconstruction error of CS-MRI is a good approach to compensate for imperfect modeling. Starting with the standard formulation of CS-MRI in Equation \ref{eq1}, we formulate our objective function as
\begin{equation}\label{eq2}
\hat x = \mathop {\arg \min }\limits_x \left\| {{F_u}x - y} \right\|_2^2 + \alpha \left\| {x - {x_p}} \right\|_2^2,
\end{equation}
where $x_p$ is an intermediate reconstruction of the MRI. We model this intermediate reconstruction $x_p$ as the summation of a ``guidance'' image $\overline x_p$ and the error image of the reconstruction $\Delta x_p$,
\begin{equation}\label{eq3}
x_p = {\overline x_p} + \Delta x_p.
\end{equation}
Substituting this into Equation \ref{eq2}, we obtain
\begin{equation}\label{eq4}
\hat x = \mathop {\arg \min }\limits_x \left\| {{F_u}x - y} \right\|_2^2 + \alpha \left\| {x - \left( {\overline x_p  + \Delta x_p} \right)} \right\|_2^2.
\end{equation}
The guidance image $\overline x_p$ is the reconstructed MRI using any chosen CS-MRI method; thus $x_p$ can be formed using existing software prior to using our proposed method for the final reconstruction. The reconstruction error $\Delta x_p$ is between the ground truth $x$ and the reconstruction $\overline x_p$. Since we don't know this at testing time, we use training data to model this error image with a neural network ${f_\theta(\mathcal{X}) }$, where $\theta$ represents the network parameters and $\mathcal{X}$ is the input to the network. Thus, Equation \ref{eq4} can be rewritten as
\begin{equation}\label{eq5}
\hat x = \mathop {\arg \min }\limits_{x,\theta} \left\| {{F_u}x - y} \right\|_2^2 + \alpha \left\| {x - \overline x_p  - {f_\theta(\mathcal{X}) }} \right\|_2^2.
\end{equation}
For a new MRI, after obtaining the guidance image $\overline x_p$ (using a pre-existing algorithm) and the mapping $\Delta x_p = {f_\theta(\mathcal{X}) }$ (using a feed-forward neural network trained on data), the proposed framework produces the final output MRI by solving the least square problem of Equation \ref{eq5}.

\section{Deep Error Correction Network (DECN)}

Following the formulation of our CS-MRI framework above and in Figure \ref{fig1}, we turn to a more detailed discussion of the optimization procedure. We next discuss each module of the proposed Deep Error Correction Network (DECN) framework.

\subsection{Guide Module}

\begin{figure}
\begin{center}
   \subfigure[fully sampled] {\label {figure2a} \includegraphics[width=0.15\textwidth]{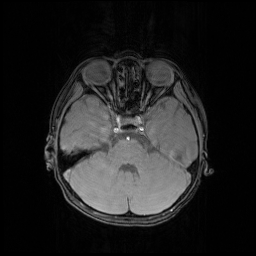}}
   \subfigure[ZF] {\label {figure2b} \includegraphics[width=0.15\textwidth]{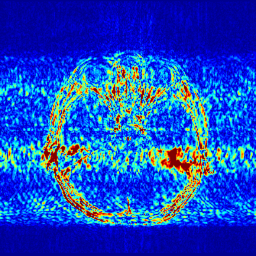}}
   \subfigure[TLMRI] {\label {figure2c} \includegraphics[width=0.15\textwidth]{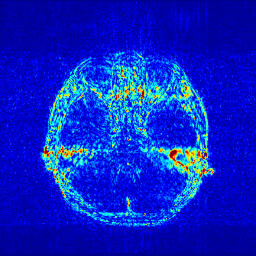}}
   \subfigure[PANO] {\label {figure2d} \includegraphics[width=0.15\textwidth]{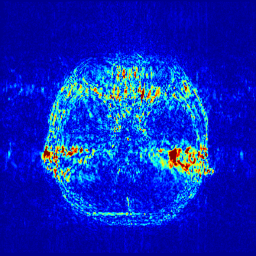}}
   \subfigure[GBRWT] {\label {figure2e} \includegraphics[width=0.15\textwidth]{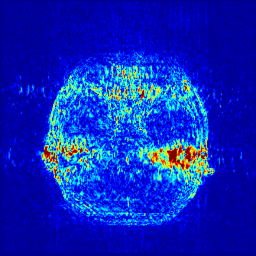}}
   \subfigure[DC-CNN] {\label {figure2f} \includegraphics[width=0.15\textwidth]{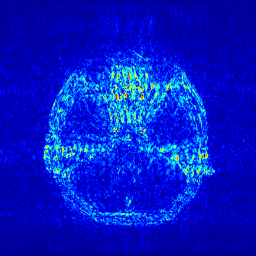}}
   \caption{The reconstruction error of a brain MRI using zero-filled, TLMRI, PANO, GBRWT and DC-CNN under 1D $30\%$ under-sampling mask.}
\label {fig2}
\end{center}
\end{figure}

With the guide module, we seek a reconstruction of the MRI $\overline x_p$ that approximates the fully-sampled MRI using a standard ``off-the-shelf'' CS-MRI approach. We denote this as
\begin{equation}\label{eq6}
\overline x_p  = {\rm{invMRI}}\left( y \right).
\end{equation}
We illustrate with reconstructions for three CS-MRI methods: TLMRI (transform learning MRI) \cite{38}, PANO (patch-based nonlocal operator) \cite{14} and GBRWT (graph-based redundant wavelet transform) \cite{9}. The PANO and GBRWT models achieve impressive reconstruction qualities because they use an nonlocal prior and adaptive graph-based wavelet transform to exploit image structures. In TLMRI, the sparsifying transform learning and the reconstruction are performed simultaneously in more efficient way than DLMRI \cite{10}. The three methods represent the state-of-the-art performance in the non-deep CS-MRI models. In Figure \ref{fig2}, we show the reconstructions error for zero-filled (itself a potential reconstruction ``algorithm''), TLMRI, PANO and GBRWT on a complexed-valued brain MRI using $30\%$ Cartesian under-sampling. The error display ranges from 0 to 0.2 with normalized data. The parameter setting will be elaborated in the Experiment Section \uppercase\expandafter{\romannumeral5}.

We also consider the deep learning DC-CNN model \cite{18} as the guide module. We also give the reconstruction error in Figure \ref{fig2}. We observe the zero-filled, TLMRI, PANO, GBRWT and DC-CNN models all suffer the structural reconstruction errors, while the DC-CNN model achieves the highest reconstruction quality with minimal errors because of its powerful model capacity. Another advantage of this CNN model is that, once the network is trained, testing is very fast compared with conventional sparse-regularization CS-MRI models. This is because no iterative algorithm needs to be run for optimization during testing since the operations are a simple feed forward function of the input. We compare the reconstruction time of TLMRI, PANO, GBRWT and DC-CNN \textit{for testing} for Figure \ref{fig2} in Table \ref{table1}.

\begin{table} [htbp]
\centering
 \caption{\label{table1} Reconstruction time of PANO, TLMRI, GBRWT and DC-CNN.}
 \begin{tabular}{lcccl}
  \toprule
                    & PANO & TLMRI & GBRWT & DC-CNN \\
  \midrule
  Runtime (seconds) & 11.37s & 127.67s & 100.60s & 0.04s \\
  \bottomrule
 \end{tabular}
\end{table}

\subsection{Error Reconstruction Module}

Using the guidance image $\overline x_p$, we can train a deep error correction module on the residual. To perform this task, we need access during training to pairs of the true, fully sampled MRI $x_{fs}$, as well as its reconstruction $\overline x_p$ found by manually undersampling the k-space of this image according to a pre-defined mask and inverting.
We then optimize the following objective function over network parameter $\theta$,
\begin{equation}\label{eq7}
\hat \theta  = \mathop {\arg \min }\limits_\theta  \frac{1}{2}\left\| {\left( {{x_{fs}} - {{\bar x}_p}} \right) - {f_\theta }\left( \mathcal{Z}(y),\overline x_p \right)} \right\|_2^2,
\end{equation}
where $\mathcal{Z}(y)$ indicates the reconstructed MRI using zero-filled and the input to the error correction module $\mathcal{X}$ is the concatenation of the zero-filled MRI $\mathcal{Z}(y)$ and the guidance MRI $\overline x_p$. Therefore, the error-correcting network is learning how to map the concatenation of the zero-filled, compressively sensed MRI and the guidance image to the residual of the true MRI using a corresponding off-the-shelf CS-MRI inversion algorithm. Now we give the rationales and explanations for the concatenation operation.

In the CS-MRI inversions, the zero-filled MR images usually serve as the starting point in the iterative optimization. Although the iterative de-aliasing can effectively remove the artifacts and achieve much more pleasing visual quality compared with zero-filled reconstruction, the distortion and information loss is inevitable in the reconstruction. To further illustrate this phenomenon, we compare the pixel-wise reconstruction errors among the zero-filling reconstruction and other reconstruction models of the MR image in Figure \ref{fig2}.

\begin{figure}
\begin{center}
   \subfigure[TLMRI $m_d$] {\includegraphics[width=0.15\textwidth]{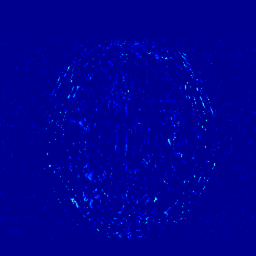}}
   \subfigure[PANO $m_d$]  {\includegraphics[width=0.15\textwidth]{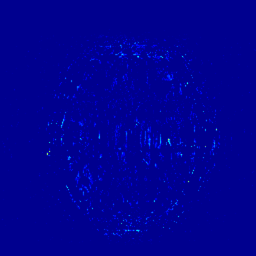}}
   \subfigure[GBRWT $m_d$] {\includegraphics[width=0.15\textwidth]{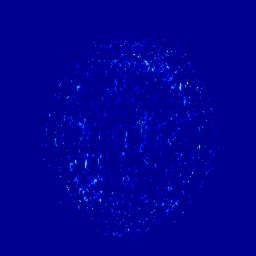}}
   \caption{The filtered difference map $d$ between the reconstruction errors of the zero-filled reconstruction and recent CS-MRI inversions.}
\label {fig22}
\end{center}
\end{figure}

We take the difference between the absolute reconstruction error of zero-filled and the compared CS-MRI methods and only keep the nonnegative values, which can be formulated as
\begin{equation}\label{eq75}
m_d = {\left( {\left| {{x_{fs}} - {\overline x_p}} \right| - \left| {{x_{fs}} -  {\cal Z}(y)} \right|} \right)_ + }.
\end{equation}
Where the operator ${\left( \cdot \right)_ + }$ set the negative values to zero. We only keep the nonnegative values in the map, which results the filtered difference map. We show the corresponding filtered difference map $m_d$ in figure \ref{fig22} in the range [0 0.2]. The bright region means the better accuracy of zero-filled reconstruction. We observe the zero-filling reconstruction provide better reconstruction accuracy on some regions, indicating the information loss in the reconstruction occurs.

To alleviate the information loss in the guide module, we introduce the concatenation operation to utilize the information from both the zero-filled MR image and guidance image as the input to the error correction network. In later Experiment Section \uppercase\expandafter{\romannumeral5}, we further validate it by the ablation study.

We again note that the network ${f_\theta }\left( \mathcal{Z}(y),\overline x_p \right)$ is paired with a particular inversion algorithm $\rm{invMRI}(y)$, since each algorithm may have unique and consistent characteristics in the errors they produce. The network ${f_\theta }\left( \mathcal{Z}(y),\overline x_p \right)$ can be any deep learning network trained using standard methods.

\subsection{Data Fidelity Module}

After the error correction network is trained, for a new undersampled k-space data $y$ for which the true $x_{fs}$ is unknown, we use its corresponding guidance image $\overline x_p = \rm{invMRI}(y)$ and the approximated reconstructed error ${f_{\theta}}\left( {\mathcal{Z}(y),\overline x_p} \right)$ to optimize the data fidelity module by solving the following optimization problem
\begin{equation}\label{eq8}
\hat x = \mathop {\arg \min }\limits_x \left\| {{F_u}x - y} \right\|_2^2 + \alpha \left\| {x - \left( {{{\bar x}_p} + {f_{ \theta }}\left( {\mathcal{Z}(y)},\overline x_p \right)} \right)} \right\|_2^2.
\end{equation}
The data fidelity module is utilized in our proposed DECN framework to correct the reconstruction by enforcing greater agreement at the sampled k-space locations. Using the properties of the fast Fourier transform (FFT), we can simplify the optimization by working in the Fourier domain using the common technique described in, e.g., \cite{11}. The optimal values for $\hat x$ in k-space can be found point-wise. This yields the closed-form solution
\begin{equation}\label{eq9}
\hat x = {F^H}\frac{{FF_u^Hy + \alpha F\left( {{{\overline x }_p} + {f_\theta }\left({\mathcal{Z}(y),\overline x_p} \right)} \right)}}{{FF_u^H{F_u}{F^H} + \alpha I}}.
\end{equation}
The regularization parameter $\alpha$ is usually set very small in the noise-free environment. We found that $\alpha = 5e{-5}$ worked well in our low-noise experiments.

The proposed DECN model can effectively exploiting the reconstruction residues. In later Experiment Section \uppercase\expandafter{\romannumeral5}, we further validate the network architecture via ablation study on the error correction strategies.

\section{Experiments}

\begin{figure*}
\begin{center}
    \subfigure[The DECN model without input concatenation and error correction (DECN-NIC-NEC)] {\label{fig33a} \includegraphics[width=1\textwidth]{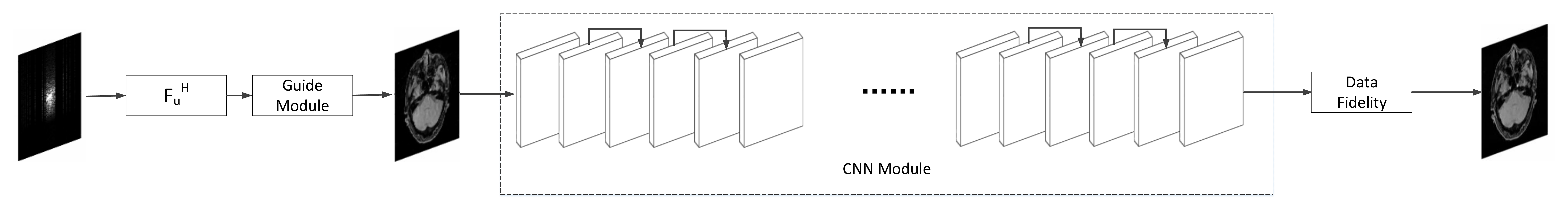}}
    \subfigure[The DECN model with input concatenation and without error correction (DECN-IC-NEC)] {\label{fig33b} \includegraphics[width=1\textwidth]{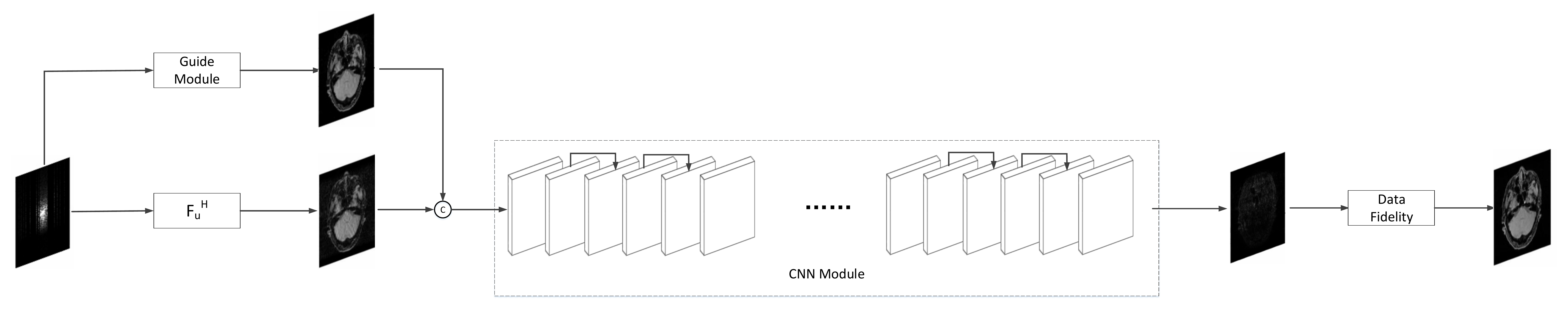}}
    \subfigure[The DECN model without input concatenation and with error correction (DECN-NIC-EC)] {\label{fig33c} \includegraphics[width=1\textwidth]{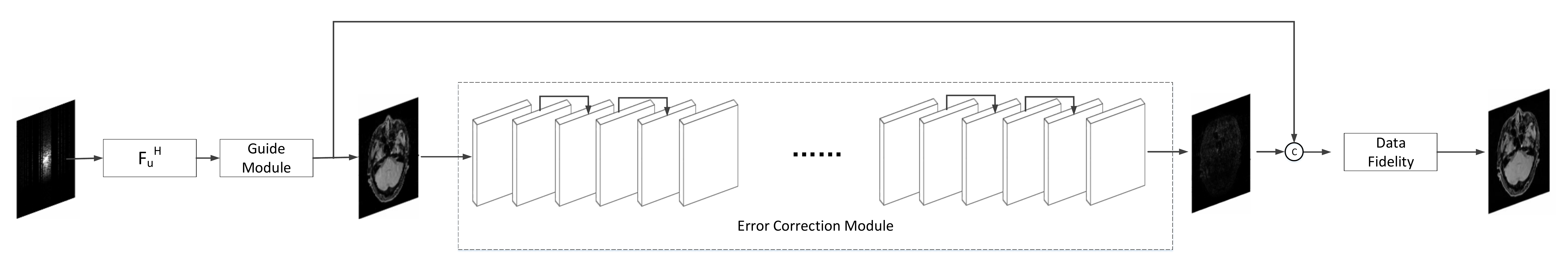}}
    \caption{The compared baseline network architectures for the ablation study to evaluate the input concatenation and error correction strategies.}
\label {fig33}
\end{center}
\end{figure*}

In this section, we present experimental results using complex-valued MRI datasets. The T1 weighted MRI dataset (size $256\times256$) is acquired on 40 volunteers with total 3800 MR images at Siemens 3.0T scanner with 12 coils using the fast low angle shot (FLASH) sequence (TR/TE = $55/3.6ms$, 220$mm^2$ field of view, $1.5mm$ slice thickness). The SENSE reconstruction is introduced to compose the gold standard full k-space, which is used to emulate the single-channel MRI. The details can be found in \cite{7}. We randomly select 80$\%$ MR images as training set and 20$\%$ MR images as testing set. Informed consent was obtained from the imaging subject in compliance with the Institutional Review Board policy. The magnitude of the full-sampled MR image is normalized to unity by dot dividing the image by its largest pixel magnitude.

Under-sampled k-space measurements are manually obtained via Cartesian and Random sampling mask with random phase encodes. Different undersampling ratios are adopted in the experiments.
\subsection{Network Architecture}


\begin{table*}[]
\centering
\caption{The objective evalution on the regular CS-MRI inversions and their DECN frameworks.}
\label{Object}
\begin{tabular}{|c|c|c|c|c|c|c|c|c|c|c|c|c|}
\hline
Sampling Pattern & \multicolumn{6}{c|}{Cartesian Under-sampling}                                     & \multicolumn{6}{c|}{Random Under-sampling}                                        \\ \hline
Sampling Ratio   & \multicolumn{2}{c|}{20\%} & \multicolumn{2}{c|}{30\%} & \multicolumn{2}{c|}{40\%} & \multicolumn{2}{c|}{20\%} & \multicolumn{2}{c|}{30\%} & \multicolumn{2}{c|}{40\%} \\ \hline
Evaluation Index & PSNR        & SSIM        & PSNR        & SSIM        & PSNR        & SSIM        & PSNR        & SSIM        & PSNR        & SSIM        & PSNR        & SSIM        \\ \hline
TLMRI            & 31.27       & 0.864       & 32.86       & 0.868       & 35.99       & 0.896       & 35.13       & 0.878       & 36.46       & 0.882       & 37.26       & 0.891       \\ \hline
PANO             & 30.71       & 0.858       & 32.65       & 0.889       & 37.40       & 0.940       & 36.94       & 0.931       & 39.36       & 0.949       & 40.74       & 0.957       \\ \hline
GBRWT            & 30.61       & 0.853       & 32.27       & 0.879       & 37.19       & 0.932       & 36.81       & 0.908       & 39.16       & 0.932       & 40.72       & 0.944       \\ \hline
DC-CNN           & 32.58       & 0.885       & 34.67       & 0.905       & 39.52       & 0.955       & 38.54       & 0.937       & 40.91       & 0.953       & 42.47       & 0.961       \\ \hline
TLMRI-DECN       & 32.77       & 0.876       & 34.41       & 0.891       & 38.62       & 0.944       & 37.60       & 0.930       & 39.54       & 0.944       & 40.72       & 0.949       \\ \hline
PANO-DECN        & 32.57       & 0.864       & 34.43       & 0.891       & 39.27       & 0.953       & 38.51       & 0.940       & 40.88       & 0.956       & 42.42       & 0.963       \\ \hline
GBRWT-DECN       & 32.58       & 0.869       & 34.41       & 0.891       & 39.07       & 0.950       & 38.48       & 0.940       & 40.79       & 0.955       & 42.36       & 0.963       \\ \hline
DC-CNN-DECN      & 33.06       & 0.898       & 35.34       & 0.922       & 39.92       & 0.956       & 38.86       & 0.939       & 41.06       & 0.954       & 42.58       & 0.962       \\ \hline
$\Delta$ TLMRI      & 1.50       & 0.012       & 1.55       & 0.023       & 2.63       & 0.048       & 2.47       & 0.052       & 3.08       & 0.062       & 3.46       & 0.068       \\ \hline
$\Delta$ PANO      & 1.86        & 0.006       & 1.78       & 0.002       & 1.87       & 0.012       & 1.57       & 0.010       & 1.52       & 0.006       & 1.68       & 0.006       \\ \hline
$\Delta$ GBRWT      & 1.97       & 0.016       & 2.14       & 0.012       & 1.88       & 0.018       & 1.67       & 0.032       & 1.63       & 0.023       & 1.64       & 0.019       \\ \hline
$\Delta$ DC-CNN      & 0.48       & 0.013       & 0.67       & 0.017       & 0.40       & 0.010       & 0.32       & 0.002       & 0.15       & 0.001       & 0.11       & 0.010       \\ \hline
\end{tabular}
\end{table*}

For the deep guide module (i.e., learning $\overline x_p$), we use the CNN architecture called deep cascade CNN (DC-CNN) \cite{18}, where the non-adjustable data fidelity layer is also incorporated into the model. This guide module consists of four blocks. Each block is formed by four consecutive convolutional layers with a shortcut and a data fidelity layer. For each convolutional layer, except the last one within a block, there are total of 64 feature maps. We use ReLU \cite{31} as the activation function.

For the error correction module (i.e., learning $f_{\theta}(\mathcal{Z}(y),\overline x_p)$), we adopt the network architecture shown in Figure \ref{fig1}. There are 18 convolutional layers with a skip layer connection as proposed in \cite{32,33} to alleviate the gradient vanish problem. We again adopt ReLU as the activation function, except for the last layer where the identity function is used to allow negative values. All convolution filters are set to $3 \times 3$ with stride set to $1$.

\subsection{Experimental Setup}
We train and test the two deep algorithms using Tensorflow \cite{34} for the Python environment on a NVIDIA GeForce GTX 1080 with 8GB GPU memory. Padding is applied to keep the size of features the same. We use the Xavier method \cite{35} to initialize the network parameters, and we apply ADAM \cite{36} with momentum. The implementation uses the initialized learning rate $0.0001$, first-order momentum $0.9$ and second momentum $0.99$. The weight decay regularization parameter is set to $0.0005$. The size of training batch is $4$. We report our performance after $20000$ training iteration of DC-CNN guide module and $40000$ iterations of error correction module.

In the guidance module, we implement the state-of-the-art CS-MRI models with the following parameter settings. In TLMRI \cite{38}, we set the data fidelity parameter $1e6/(256\times256)$, the patch size $36$, the number of training signals $256\times256$, the sparsity fraction $4.6\%$, the weight on the negative log-determinat+Frobenius norm terms $0.2$, the patch overlap stride $1$, the DCT matrix is used as initial transform operator, the iterations $50$ times for optimization. The above parameter setting follows the advices from the author \cite{38}. In PANO \cite{14}, we use the implementation with parallel computation provided by \cite{14}. The data fidelity parameter is set $1e6$ with zero-filled MR image as initial reference image. The non-local operation is implemented twice to yield the MRI reconstruction. In GBRWT \cite{9}, we set the data fidelity parameter $5\times1e3$. The Daubechies redundant wavelet sparsity is used as regularization to obtain the reference image. The graph is trained $2$ times.

\subsection{Ablation Study}

To validate the architecture of the proposed DECN model, we conduct the ablation study by comparing the DECN framework with other Baseline network architectures in Figure \ref{fig33}, which we refer the model in Figure \ref{fig33a} as DECN with no input concatenation and error correction (DECN-NIC-NEC). With the guide module, a later cascaded CNN module learns the mapping from the pre-reconstructed MR image to the full-sampled MR image. Likewise, we name the models in Figure \ref{fig33b} (DECN-IC-NEC) and Figure \ref{fig33c} (DECN-NIC-EC). By comparing the DECN-NIC-NEC framework with the DECN-IC-NEC framework, we evaluate the benefit brought by the concatenating the zero-filled MR images and corresponding guide MR images as the input to compensate the information loss in the guide module. In Figure \ref{fig22}, we give the illustration the information from zero-filled MR images and guide images can be shared. By comparing the DECN-NIC-NEC framework with the DECN-NIC-EC framework, we evaluate how the error correction strategy improves the reconstruction accuracy compared with simple cascade manner.

\begin{figure}
\begin{center}
    {\includegraphics[width=0.49\textwidth]{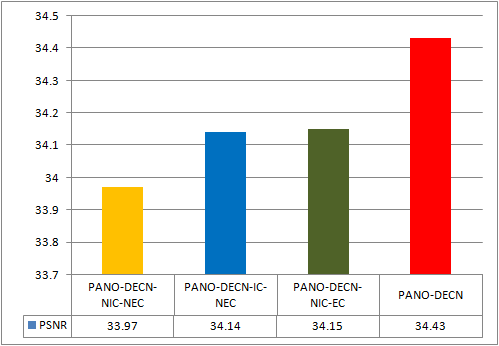}}
    \caption{The PSNR (top) and SSIM (bottom) comparison of the regular CS-MRI methods (solid line) and their corresponding DECN-based methods (dashed line). All the tested data are shown in the comparison. The x-axis shows the index of the tested data and the y-axis shows the model performances.}
\label {fig44}
\end{center}
\end{figure}

We conduct the experiments using PANO with the Cartesian under-sampling mask shown in Figure \ref{fig4} as the guide module. We give the averaged PSNR (peak signal-to-noise ratio) results in Figure \ref{fig44}. We observe the PANO-DECN-IC-NEC and PANO-DECN-NIC-EC both outperforms the PANO-DECN-NIC-NEC with the similar margins about 0.2dB in PSNR. While the proposed PANO-DECN model with the input concatenation and error correction outperforms the PANO-DECN-NIC-NEC about 0.5 dB in PSNR. The ablation study shows the input concatenation and error correction strategies can effectively improve the model performance in the DECN framework.

\subsection{Results}

\begin{figure}[htb!]
\begin{center}
   \subfigure[\tiny Fully-sampled] {\includegraphics[width=0.115\textwidth]{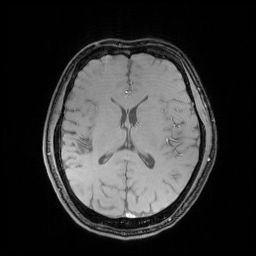}}
   \subfigure[\tiny 1D $30\%$ Mask] {\label {fig4b} \includegraphics[width=0.115\textwidth]{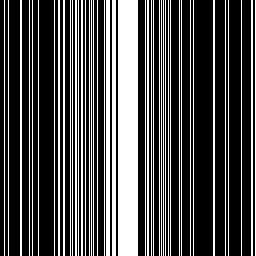}}
   \subfigure[\tiny Zero-filled] {\includegraphics[width=0.115\textwidth]{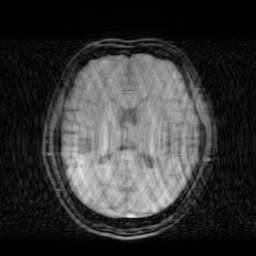}}\\
   \subfigure[\tiny Full-sampled] {\includegraphics[width=0.115\textwidth]{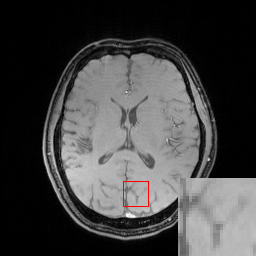}}
   \subfigure[\tiny Full-sampled] {\includegraphics[width=0.115\textwidth]{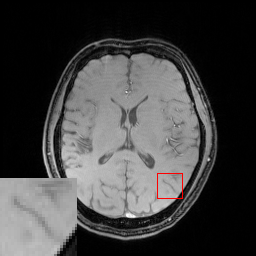}}
   \subfigure[\tiny Full-sampled] {\includegraphics[width=0.115\textwidth]{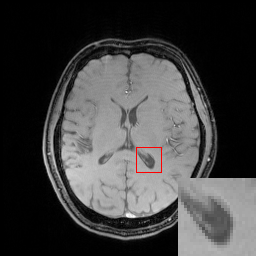}}
   \subfigure[\tiny Full-sampled] {\includegraphics[width=0.115\textwidth]{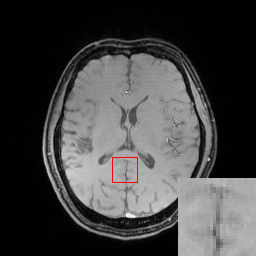}}\\
   \subfigure[\tiny TLMRI] {\includegraphics[width=0.115\textwidth]{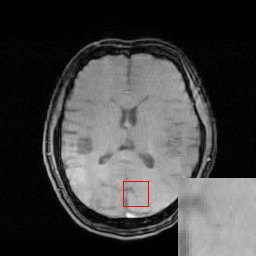}}
   \subfigure[\tiny PANO] {\includegraphics[width=0.115\textwidth]{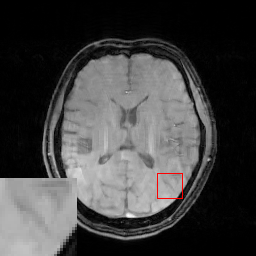}}
   \subfigure[\tiny GBRWT] {\includegraphics[width=0.115\textwidth]{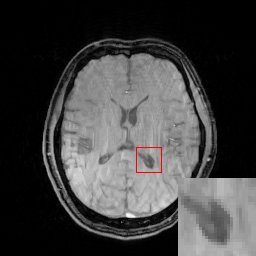}}
   \subfigure[\tiny DC-CNN] {\includegraphics[width=0.115\textwidth]{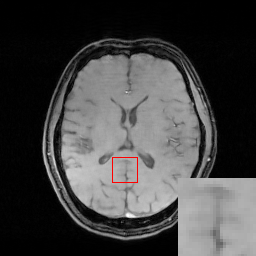}}\\
   \subfigure[\tiny TLMRI-DECN] {\includegraphics[width=0.115\textwidth]{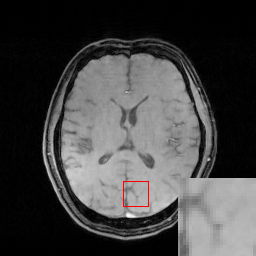}}
   \subfigure[\tiny PANO-DECN] {\includegraphics[width=0.115\textwidth]{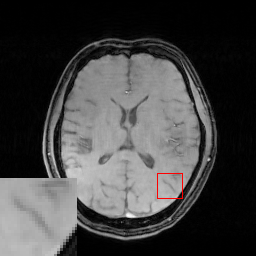}}
   \subfigure[\tiny GBRWT-DECN] {\includegraphics[width=0.115\textwidth]{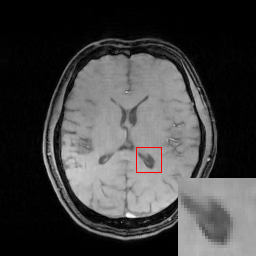}}
   \subfigure[\tiny DC-CNN-DECN] {\includegraphics[width=0.115\textwidth]{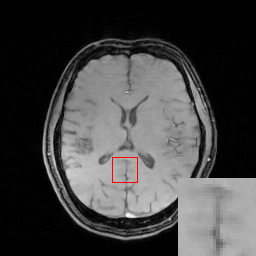}}\\
   \subfigure[\tiny $\Delta$ TLMRI] {\includegraphics[width=0.115\textwidth]{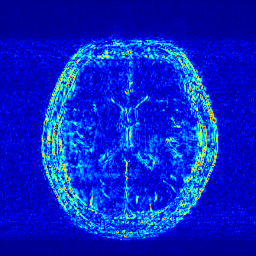}}
   \subfigure[\tiny $\Delta$ PANO] {\includegraphics[width=0.115\textwidth]{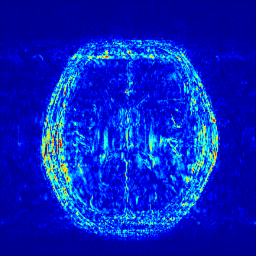}}
   \subfigure[\tiny $\Delta$ GBRWT] {\includegraphics[width=0.115\textwidth]{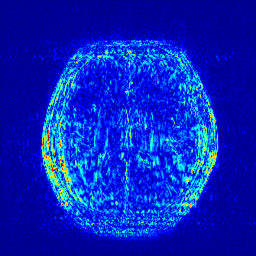}}
   \subfigure[\tiny $\Delta$ DC-CNN] {\includegraphics[width=0.115\textwidth]{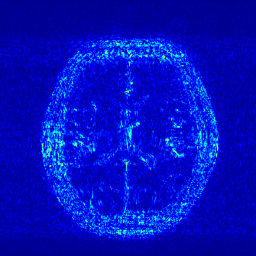}}\\
   \subfigure[\tiny $\Delta$ TLMRI-DECN] {\includegraphics[width=0.115\textwidth]{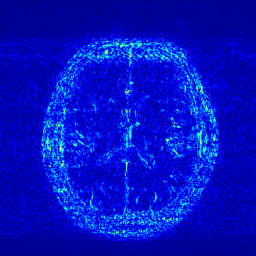}}
   \subfigure[\tiny $\Delta$ PANO-DECN] {\includegraphics[width=0.115\textwidth]{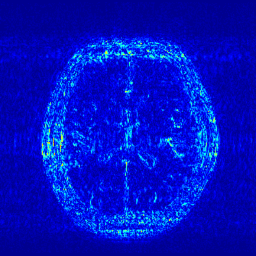}}
   \subfigure[\tiny $\Delta$ GBRWT-DECN] {\includegraphics[width=0.115\textwidth]{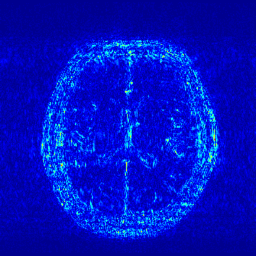}}
   \subfigure[\tiny $\Delta$ DC-CNN-DECN] {\includegraphics[width=0.115\textwidth]{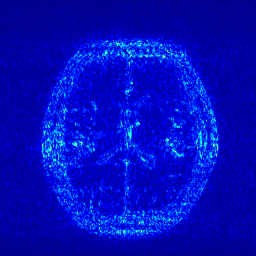}}\\
   \caption{We show the reconstruction results of our DECN model with local area magnification. We also show the reconstruction error for our DECN model under different guide module in the last row.}
\label {fig4}
\end{center}
\end{figure}

\begin{figure}[htb!]
\begin{center}
   \subfigure[\footnotesize Fully-sampled]  {\includegraphics[width=0.15\textwidth]{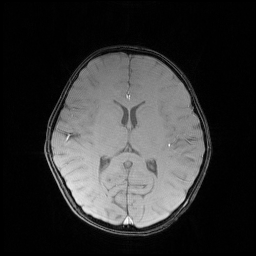}}
   \subfigure[\footnotesize 2D $20\%$ Mask] {\label {fig5b} \includegraphics[width=0.15\textwidth]{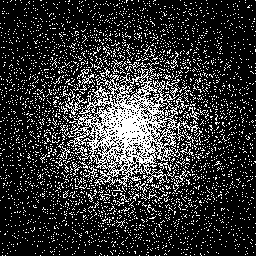}}
   \subfigure[\footnotesize Zero-filled]    {\includegraphics[width=0.15\textwidth]{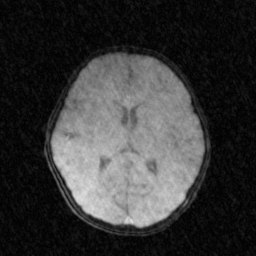}}\\
   \subfigure[\footnotesize Full-sampled]   {\includegraphics[width=0.15\textwidth]{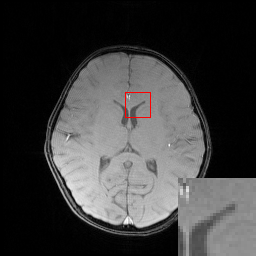}}
   \subfigure[\footnotesize Full-sampled]   {\includegraphics[width=0.15\textwidth]{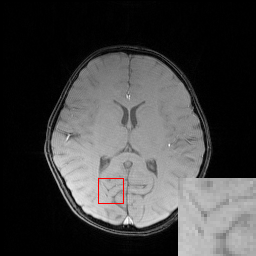}}
   \subfigure[\footnotesize Full-sampled]   {\includegraphics[width=0.15\textwidth]{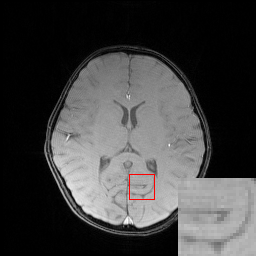}}\\
   \subfigure[\footnotesize TLMRI]   {\includegraphics[width=0.15\textwidth]{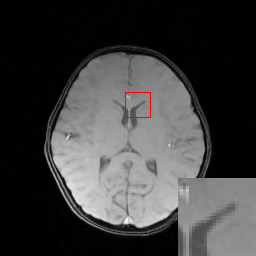}}
   \subfigure[\footnotesize PANO]   {\includegraphics[width=0.15\textwidth]{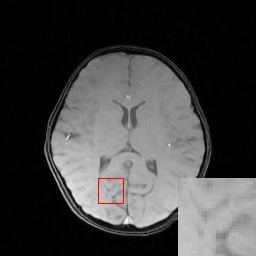}}
   \subfigure[\footnotesize GBRWT]   {\includegraphics[width=0.15\textwidth]{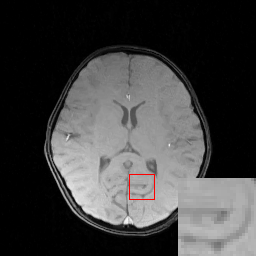}}\\
   \subfigure[\footnotesize TLMRI-DECN]   {\includegraphics[width=0.15\textwidth]{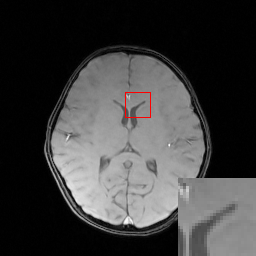}}
   \subfigure[\footnotesize PANO-DECN]   {\includegraphics[width=0.15\textwidth]{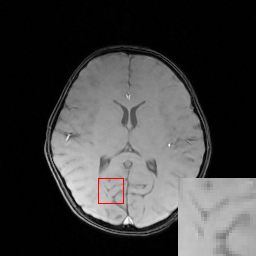}}
   \subfigure[\footnotesize GBRWT-DECN]   {\includegraphics[width=0.15\textwidth]{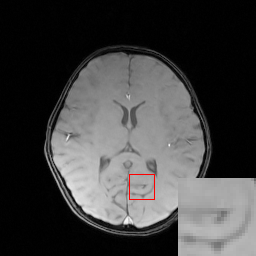}}\\
   \subfigure[\footnotesize $\Delta$ TLMRI]          {\includegraphics[width=0.15\textwidth]{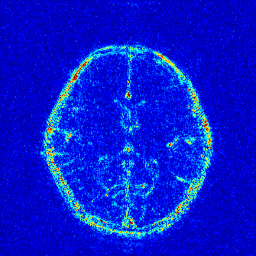}}
   \subfigure[\footnotesize $\Delta$ PANO]           {\includegraphics[width=0.15\textwidth]{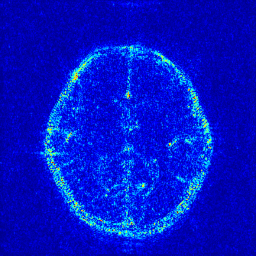}}
   \subfigure[\footnotesize $\Delta$ GBRWT]          {\includegraphics[width=0.15\textwidth]{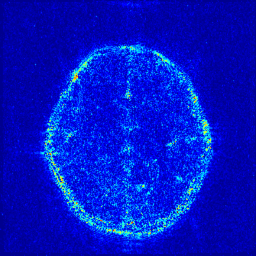}}\\
   \subfigure[\footnotesize $\Delta$ TLMRI-DECN] {\includegraphics[width=0.15\textwidth]{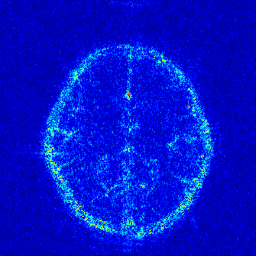}}
   \subfigure[\footnotesize $\Delta$ PANO-DECN]  {\includegraphics[width=0.15\textwidth]{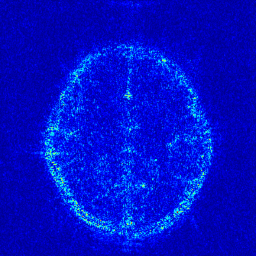}}
   \subfigure[\footnotesize $\Delta$ GBRWT-DECN] {\includegraphics[width=0.15\textwidth]{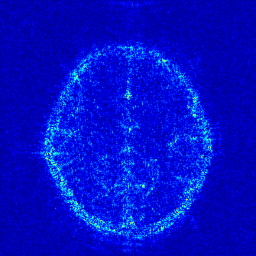}}
   \caption{We show the reconstruction results of our DECN framework on TLMRI, PANO and GBRWT methods with local area magnification on Random $20\%$ under-sampling mask. We also show the reconstruction error for our DECN model under different guide module in the last row.}
\label {fig5}
\end{center}
\end{figure}

We evaluate the proposed DECN framework using PSNR and SSIM (structural similarity index) \cite{37} as quantitative image quality assessment measures. We give the quantitative reconstruction results of all the test data on different under-sampling patterns and different under-sampling ratios in Table \ref{Object}. We show the Cartesian $30\%$ under-sampling mask in Figure \ref{fig4b} and the Random $20\%$ under-sampling mask in Figure \ref{fig5b}. We observe that DECN improved all off-the-shelf CS-MRI inversion methods on all the under-sampling patterns. Since the Random mask enjoys the more incoherence than the Cartesian mask with the same under-sampling ratio, the CS-MRI achieves better reconstruction quality on the Random masks. Also, we observe the plain DC-CNN model already achieves good reconstruction accuracy, leaving less structural errors for the error correction module, leading to the limited performance improvement about 0.1 dB on the Random $20\%$ and $30\%$ masks. While for other CS-MRI inversions on various sampling patterns, the improvements are at least 1.5dB or even up to 3.5 dB.

In Figure \ref{fig4}, we show reconstruction results and the corresponding error images of an example from the test data on the 1D $30\%$ under-sampling mask. With local magnification on the red box, we observe that by learning the error correction module, the fine details, especially the low-contrast structures are better preserved, leading to a better reconstruction.

In Figure \ref{fig5}, we also compare the MR images produced by the TLMRI, PANO and GBRWT with their DECN counterparts on the 2D $20\%$ under-sampling mask. The subjective comparison DC-CNN and DE-CNN-DECN isn't included because of the limited improvement. The results are consistent with our observation in Cartesian under-sampling case.


\section{Conclusion}
We have proposed a deep error correction framework for the CS-MRI inversion problem. Using any off-the-shelf CS-MRI algorithm to construct a template, or ``guide'' for the final reconstruction, we use a deep neural network that learns how to correct for errors that typically appear in the chosen algorithm. Experimental results show that the proposed model achieves consistently improves a variety of CS-MRI inversion techniques.

\section*{Acknowledgment}

The work is supported in part by National Natural Science Foundation of China under Grants 61571382, 81671766, 61571005, 81671674, U1605252, 61671309 and 81301278, Guangdong Natural Science Foundation under Grant 2015A030313007, Fundamental Research Funds for the Central Universities under Grant 20720160075, 20720150169, CCF-Tencent research fund, Natural Science Foundation of Fujian Province of China (No.2017J01126), and the Science and Technology funds from the Fujian Provincial Administration of Surveying, Mapping, and Geoinformation.

\bibliographystyle{ieee}
\bibliography{DECN}

\end{document}